\documentclass[conference]{IEEEtran}
\IEEEoverridecommandlockouts
\usepackage{cite}
\usepackage{amsmath,amssymb,amsfonts}
\usepackage{algorithmic}
\usepackage{graphicx}
\usepackage{textcomp}
\usepackage{xcolor}
\def\BibTeX{
    {\rm B\kern-.05em{\sc i\kern-.025em b}\kern-.08em
    T\kern-.1667em\lower.7ex\hbox{E}\kern-.125emX
    }
   
}
\begin{document}

\title{EL-VIT: Probing Vision Transformer with Interactive Visualization}

\author{

Hong Zhou, Rui Zhang, Peifeng Lai, Chaoran Guo, Yong Wang*\thanks{
*Corresponding author.\newline
• Hong Zhou, Zhida Sun and Junjie Li are with Shenzhen University, E-mail: \{hzhou$\mid$zhida.sun$\mid$jj.li\}@szu.edu.cn
\newline
• Rui Zhang, Peifeng Lai and Chaoran Guo are with Shenzhen University, E-mail: \{zhangrui2021$\mid$laipeifeng2022$\mid$guochaoran2021\}@email.szu.edu.cn
\newline
• Yong Wang is with Singapore Management University, E-mail: yongwang@smu.edu.sg
}, Zhida Sun and Junjie Li
% \IEEEauthorblockA{hzhou@szu.edu.cn}
% \and
% \IEEEauthorblockN{Rui Zhang†}
% \IEEEauthorblockA{zhangrui2021@email.szu.edu.cn}
% \and
% \IEEEauthorblockN{3\textsuperscript{rd} Peifeng Lai†}
% \IEEEauthorblockA{laipeifeng2022@email.szu.edu.cn}
% \and
% \IEEEauthorblockN{4\textsuperscript{th} Chaoran Guo†}
% \IEEEauthorblockA{guochaoran2021@163.com}
% \and
% \IEEEauthorblockN{5\textsuperscript{th} Yong Wang*\thanks{*Corresponding author.}}
% \IEEEauthorblockA{yongwang@smu.edu.sg}
% \and
% \IEEEauthorblockN{6\textsuperscript{th} Zhida Sun†}
% \IEEEauthorblockA{zhida.sun@szu.edu.can}
% \and
% \IEEEauthorblockN{7\textsuperscript{th} Junjie Li†}
% \IEEEauthorblockA{jj.li@szu.edu.cn}
% \and

% \\College of Computer Science and Software Engineering, ShenZhen University, ShenZhen, China†
% \\The School of Computing and Information Systems, Singapore Management University, Singapore, Singapore*
}

\maketitle

\begin{figure*}[h]
\centerline{\includegraphics[width=\textwidth]{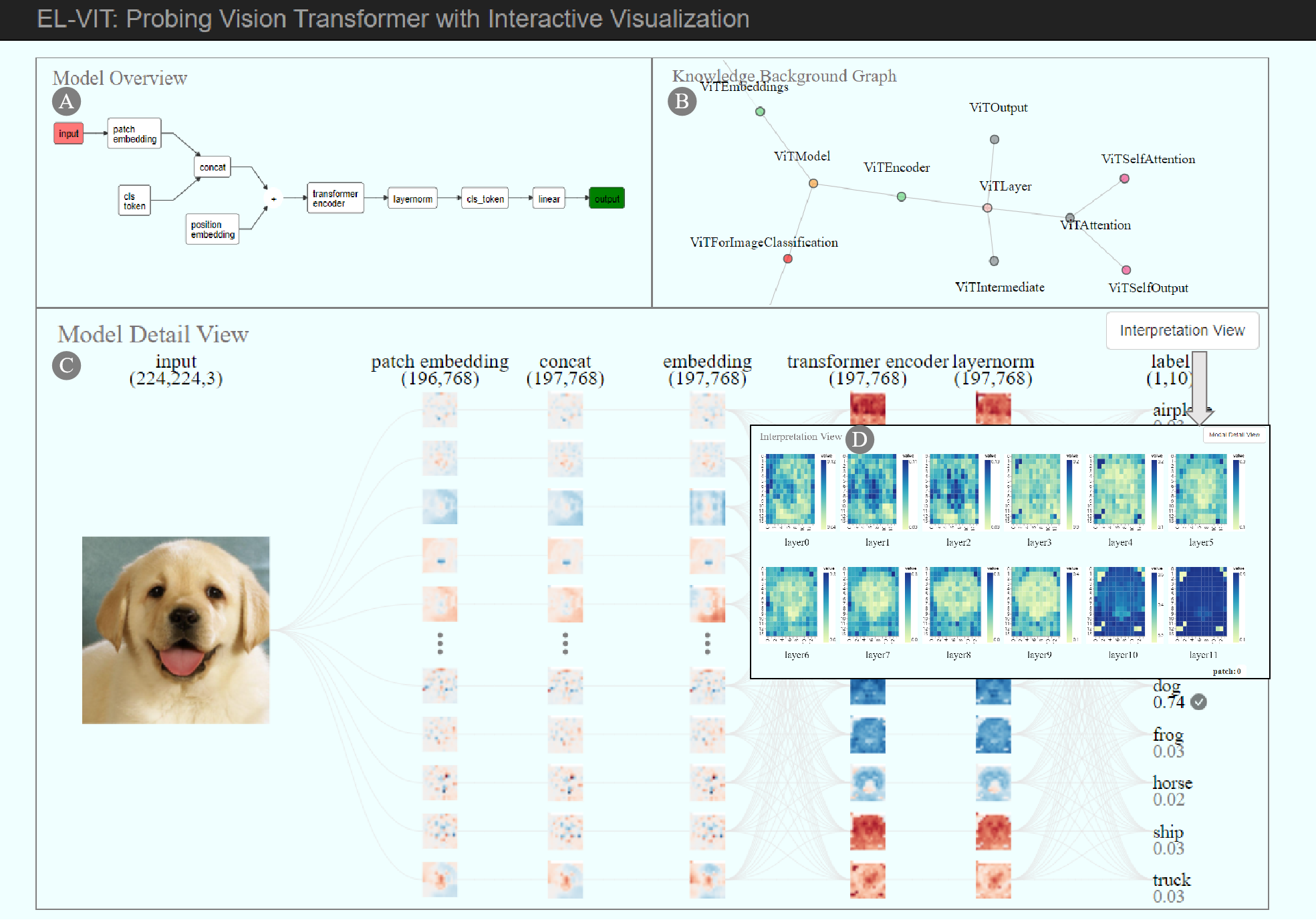}}
\caption{The overview diagram of the visualization system.}
\label{fig1}
\end{figure*}

\begin{abstract}
Nowadays, Vision Transformer (ViT) is widely utilized in various computer vision tasks, owing to its unique self-attention mechanism. However, the model architecture of ViT is complex and often challenging to comprehend, leading to a steep learning curve. ViT developers and users frequently encounter difficulties in interpreting its inner workings. Therefore, a visualization system is needed to assist ViT users in understanding its functionality. This paper introduces EL-VIT, an interactive visual analytics system designed to probe the Vision Transformer and facilitate a better understanding of its operations. The system consists of four layers of visualization views. The first three layers include model overview, knowledge background graph, and model detail view. These three layers elucidate the operation process of ViT from three perspectives: the overall model architecture, detailed explanation, and mathematical operations, enabling users to understand the underlying principles and the transition process between layers. The fourth interpretation view helps ViT users and experts gain a deeper understanding by calculating the cosine similarity between patches. Our two usage scenarios demonstrate the effectiveness and usability of EL-VIT in helping ViT users understand the working mechanism of ViT.
\end{abstract}


\section{Introduction}
Deep learning has become an integral part of our daily lives, with its application spanning across various industries. While deep learning models have continuously improved in terms of performance, they have also grown in complexity, which raises the learning curve. This paper focuses on the Vision Transformer (ViT) model, a deep learning architecture that incorporates principles from natural language processing into computer vision. ViT\cite{b1} has demonstrated its superiority over Convolutional Neural Networks (CNN) in various computer vision downstream tasks, including image classification, object detection, semantic segmentation, and image captioning. Consequently, the ViT model has generated significant interest among practitioners, students, and experts, prompting many individuals  to explore this technology.  However, for novices, understanding the inner workings of the ViT model can be challenging, given its intricate layer structures, inter-layer data transmission methods, convolution, and multi-head attention mechanisms. Taking image classification as an example, beginners may struggle to comprehend the journey from initial image data to the ultimate classification prediction. Similarly, computer vision students may feel difficult to understand concepts such as Query, Key, and Value, and experts may also be perplexed by the inner workings of ViT and the principles underlying its classification. Therefore, the development of a visualization system to facilitate the understanding and application of this technology for both beginners and experts is important.

To ensure a positive user experience with the visualization system, we encountered several challenges during the design of the visualization tool. Firstly, the layer structure of ViT is highly intricate, accompanied by a vast number of parameters. It not only possesses a deep layer structure (e.g., the simplest ViT model, ViT-B/16, comprises 12 Transformer blocks), but also incorporates various types of layers (such as Query, Key, Value, Glue, etc.), each involving distinct mathematical operations. Comprehensive guidance is necessary to assist users in gaining a thorough understanding of the ViT model. Secondly, existing visualization tools lack a unified overview of educational aids and interpretability explanations. Nowadays, numerous visualization systems employ interactive visualizations to assist users in exploring and learning deep learning models, such as CNN Explainer\cite{b2} and DNN Genealogy\cite{b3}, which respectively aid beginners in understanding convolutional neural networks and deep neural networks. Additionally, some visualization systems are dedicated to providing interpretability explanations for deep learning models, aiding experts in understanding and diagnosing these models \cite{b4,b5}. Both aspects are crucial, and EL-VIT is proposed to address these challenges with the aim of facilitating the understanding and learning of the ViT model for both experts and non-experts. 

The major contributions of this paper are as follows:
\begin{itemize}
\item \textbf{We introduce an interactive visualization tool, EL-VIT, to help ViT learners and users to explore ViT models.} Currently, many visualization tools cater to experts, with limited availability of educational-focused visualization tools. The proposed visualization platform in this paper aims to fill this research gap and serves educational purposes.
\item \textbf{We utilize a multi-view visualization system design, each view providing unique insights into the model.} This approach fosters a holistic comprehension of the ViT model and promotes the acquisition of model-related knowledge. EL-VIT is web-based, requiring no backend programming or the need for users to install specific software. Users can explore the operations of ViT simply by accessing the web page. EL-VIT also incorporates various interactive visualization methods, providing users with a seamless experience and aiding in their better comprehension of ViT.
\item \textbf{We offer an innovative approach to enhancing the interpretability of the ViT model.} Rather than concentrating on visualizing attention weights, EL-VIT calculates cosine similarity for the outputs of each Transformer block. It reveals that patches corresponding to the same object tend to exhibit higher similarity, and the token used for classification, CLS token, often demonstrates greater similarity to the patches associated with the classified objects. By providing interpretation view, EL-VIT facilitates convenient model exploration for users.
\end{itemize}

\section{Background}

Within this section, we present a concise retrospective analysis of the Vision Transformer model, delivering a succinct exposition of the foundational principles and technical specifics of the ViT model. This segment lays the groundwork for the subsequent visualization chapter of this paper.

\begin{figure}[t]
  \centerline{\includegraphics[width=0.5\textwidth]{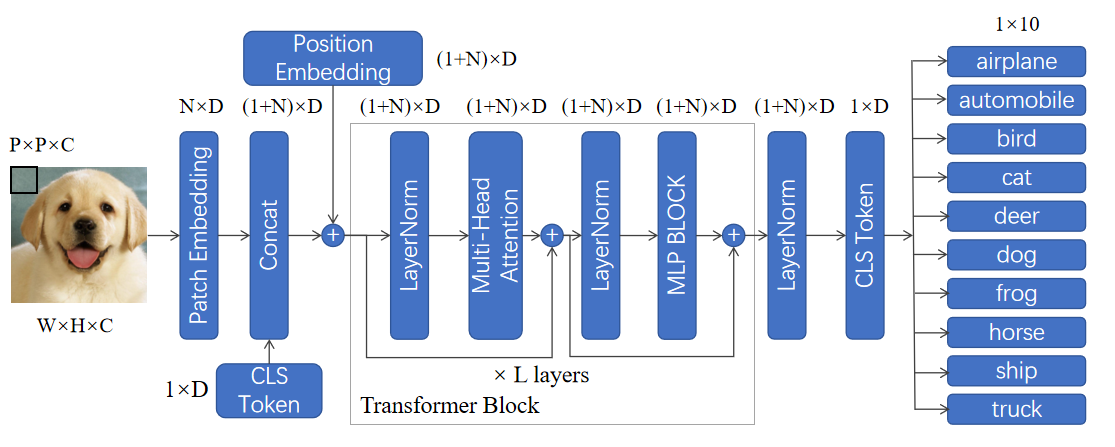}}
  \caption{The architectural diagram of the ViT-B/16 model.}
  \label{fig2}
\end{figure}

Transformer\cite{b6} has achieved significant success in the field of natural language processing, leading to its application in computer vision. On large-scale datasets, ViT has outperformed CNN and demonstrated favorable transferability in downstream tasks. Specifically, as illustrated in Fig.~\ref{fig2}, within the ViT model, when provided with an input image $X \in R^{H\times W\times C}$, where H, W, and C represent the image's height, width, and number of channels, respectively, the image is divided into N patches. Any patch's shape is $P\times P\times C$ with $N=HW/P^2$. The number of convolutional kernels is denoted as D. Subsequently, a linear transformation is performed using D convolutional kernels, each of shape is $P\times P\times C$, applied to N patches, resulting in a two-dimensional matrix. This matrix is concatenated with a CLS token, and position encoding is added. This process yields the embedding layer, as illustrated in Equation \ref{eq1}. $z_0$ precisely corresponds to the input of the Transformer encoder. The Transformer encoder consists of L layers of Transformer blocks, each of which comprises various distinct layers. These layers encompass LayerNorm, Multi-Head Self-Attention (MSA), and Multi-Layer Perceptron (MLP) blocks. Prior to each block, LayerNorm (LN) is applied, and after each block, residual connections are introduced, as explicitly defined in Equations \ref{eq2} and \ref{eq3}. Following these steps, the results are extracted with 0 dimension, and the CLS token is obtained. Subsequently, layer normalization, as demonstrated in Equation \ref{eq4}, yields the output y, where $z^0_L$ represents the CLS token. The dimensionality of y is 1×D. Finally, classification predictions are obtained through a linear transformation.

\begin{equation}
z_0=[x_{cls};x^1_pE;x^2_pE;...;x^N_pE]+E_{pos},  E_{pos}\in R^{(N+1)\times D} \label{eq1}
\end{equation}

\begin{equation}
z_l^\prime =MSA(LN(z_{l-1}))+z_{l-1}, l=1...L \label{eq2}
\end{equation}

\begin{equation}
z_l =MLP(LN(z_l^\prime))+z_l^\prime, l=1...L \label{eq3}
\end{equation}

\begin{equation}
y=LN(z^0_L) \label{eq4}
\end{equation}

Within the ViT forward process, the attention mechanism assumes a crucial role. As outlined in Fig.~\ref{fig2}, it resides within the multi-head attention layer of the Transformer block. Following normalization in the preceding layer, the resultant matrices are subject to separate multiplications with $W^Q$, $W^K$, and $W^V$, yielding Q, K, and V. Consequently, self-attention can be computed:

\begin{equation}
Attention(Q, K, V)=A\cdot V=softmax(\frac{Q\cdot K^T}{\sqrt{d_k}})\cdot V \label{eq5}
\end{equation}

Within this paper, $A\in R^{N\times N}$ denotes the attention weights, which determine the significance of relationships between different positions in the input sequence. Given that this paper employs the ViT-B/16 model, it follows that H = W = 224, C = 3, P = 16, D = 768, and N = 196.

\section{Related Work}

\subsection{Visualization for Deep Learning Education}\label{subsec1}
As researchers continually develop visualization systems, the number of visualization systems designed for educational purposes has gradually increased. These systems aim to assist beginners in better understanding concepts related to deep learning models. For example, the web-based MNIST demo on ConvNetJS\cite{b7} dynamically illustrates the training process and intermediate results of digit recognition. Teachable Machine\cite{b8} is a web-based GUI tool that enables users to learn about machine learning classification by creating and using their own classification models. DNN Genealogy\cite{b3} serves as an educational tool that provides visual analytical guidance for understanding and applying deep neural networks through interviews, surveys, and summarization and analysis of 66 representative DNNs. TensorFlow Playground\cite{b9} allows users to experiment with neural networks through direct manipulation, making it accessible to users unfamiliar with deep learning. GAN Lab\cite{b10} enables interactive training of generative models for learning and experimenting with adversarial neural networks. CNN Explainer\cite{b2} uses real image data to explain the model structure and mathematical operations of CNNs, addressing common issues encountered by beginners learning CNNs through interactive visualization. Furthermore, interactive online scientific journals like Distill\cite{b11,b12} encourage us to move beyond traditional academic formats, making interactive articles with interactive visualizations explaining deep learning models increasingly popular.

It is noteworthy that EL-VIT is developed based on CNN Explainer. However, for comparative purposes, our visualization system offers more advanced features, with two key distinctions. Firstly, as an educational tool, the inclusion of a systematic architectural diagram of the model is essential, particularly for complex models. This allows users to grasp the entire process from input to output comprehensively. Additionally, the system incorporates explanations of terminology and source code to assist users in learning relevant concepts. Secondly, EL-VIT provides an alternative approach to enhancing the interpretability of the ViT model. This allows users not only to acquire knowledge about ViT but also to gain insights into ViT's classification decisions.

\subsection{Visual Analysis for Deep Learning Model}\label{subsec2}
As the field of deep learning visualization continues to evolve, in addition to serving as educational tools, there are numerous visualization systems designed to assist experts in analyzing their models and predictions, explaining how models respond to their datasets. For example, CNNVis\cite{b4} is employed to aid experts in comprehending, diagnosing, and refining deep CNN models. LSTMVis\cite{b13} and RNNVis\cite{b14} provide tools for users to explore the hidden states of RNNs. DeepNLPVis\cite{b5} effectively identifies and diagnoses issues in deep NLP models used for text classification. GNNExplainer\cite{b15} offers interpretability for predictions of any GNN-based model in any graph-based machine learning task. CorGIE\cite{b16} is designed to check whether Graph Neural Networks (GNNs) learn significant features from graphs, and GNNLens\cite{b31} facilitate the prediction error analysis of GNNs. Some tools allow users to diagnose models during training. For example, DeepEyes\cite{b17} progressively analyzes DNNs during training by sampling subregions of the input space. DGMTracker\cite{b18} enables experts to diagnose and monitor the training process of generative models by visualizing time-series data on data flow graphs. GANViz\cite{b19} offers a comprehensive understanding of the adversarial process of GAN models, helping experts assess and explain training results through multiple views.

The aforementioned work has achieved significant success in the field of deep learning visualization. Additionally, there are several interpretability methods to help us understand Transformer models. For instance, interpretability has been achieved by employing attention mechanisms to explain the model\cite{b20,b21,b22,b23}. Attribution through self-attention has been used to elucidate the internal information interactions of the Transformer, aiding in the identification of critical attention heads\cite{b24}. The allocation of local correlations through deep Taylor decomposition principles, followed by the propagation of these correlations across layers, has been employed\cite{b25}. Visual explanations have been provided by assessing the similarity between two regions of images\cite{b26}. In comparison, our approach introduces a novel perspective by analyzing only the output obtained from a single image through Transformer blocks, specifically by comparing the cosine similarity between any two patches of the output.

\section{Design Goal}\label{sec4}

EL-VIT endeavors to create a visual platform for users to learn about and explore ViT. Nonetheless, diverse user types possess distinct demands for visualization systems. Consequently, we seek to have the visualization system accomplish the following goals:

\begin{figure}[t]
  \centering
  \includegraphics[width=0.5\textwidth]{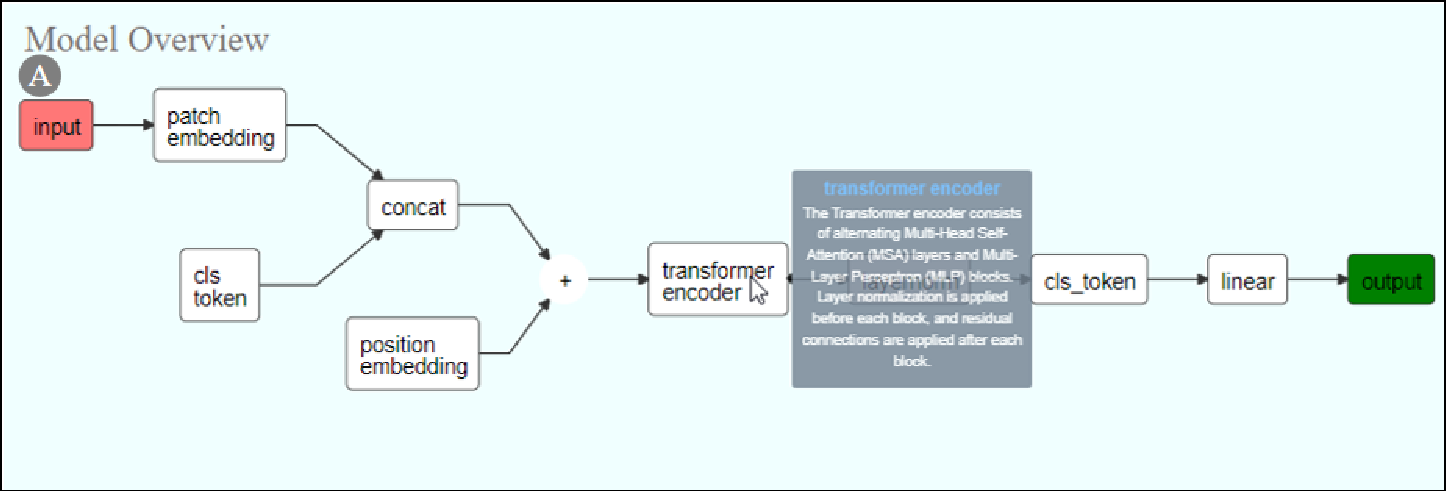}
  \includegraphics[width=0.5\textwidth]{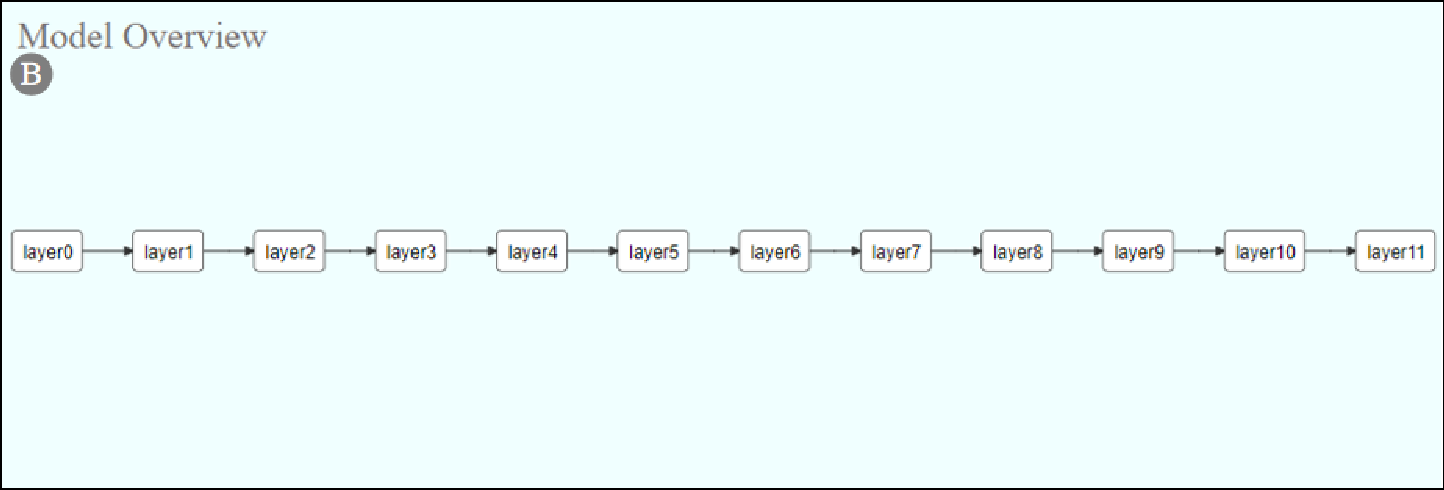}
  \includegraphics[width=0.5\textwidth]{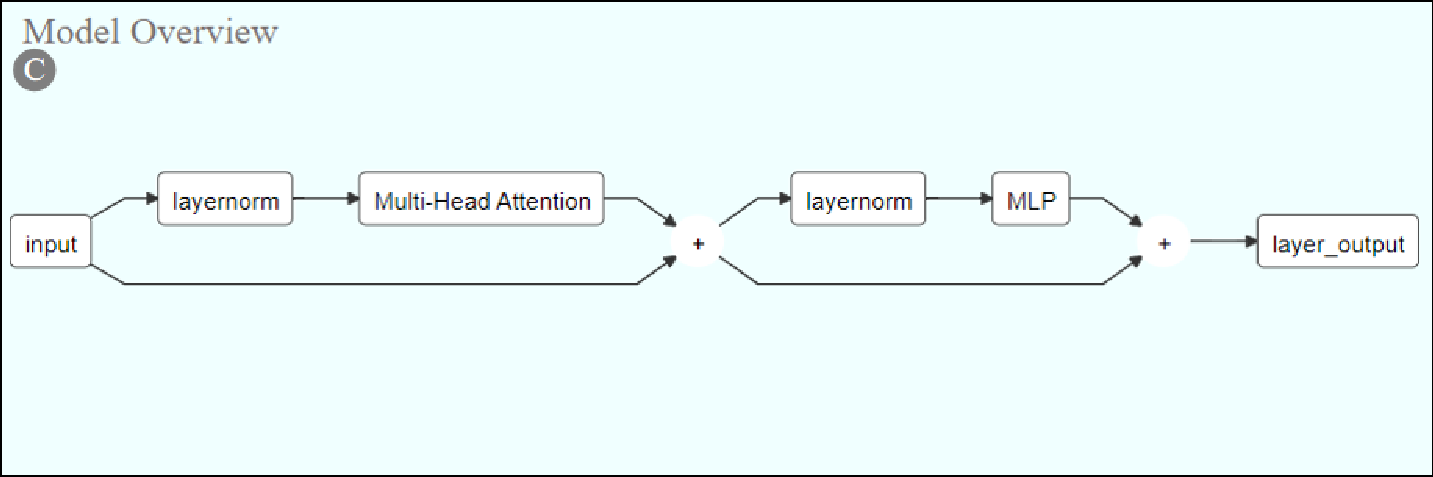}
  \caption{Model Overview. (A) illustrates the overall structure of the ViT model. (B) showcases the 12 layers of the Transformer encoder. (C) details the composition of each Transformer block at every layer.}
  \label{fig3}
\end{figure}

\begin{itemize}
\item \textbf{Catering to both global and detailed exploration of the model simultaneously.} The ViT model possesses a deep layer structure, for example, in the case of the relatively simpler ViT-B/16 model, it consists of three modules: the Embedding layer, Transformer Encoder, and MLP Head. In particular, the Transformer Encoder consists of 12 layers of Transformer blocks, with each block comprising LayerNorm, MLP, and multi-head attention components. Furthermore, the multi-head attention itself encompasses various types of layer structures. Hence, there is a need for a dual approach encompassing both global and detailed exploration, ensuring that users do not feel lost while navigating the system.
\item \textbf{Illustrating the complete data flow process of the ViT model.} The model structure of ViT is highly intricate, with various types of layers involving distinct mathematical operations. Users often find it challenging to comprehend the transformations occurring between these layers. Given these considerations, EL-VIT aims to provide a comprehensive data flow process, aiding users in understanding how image data undergoes step-by-step transformations to yield classification predictions.
\item \textbf{Achieving the interpretability of ViT.} Certain users aim to gain insight into both the practical application and the underlying reasons for the effectiveness of ViT. We propose an innovative approach that involves visualizing the cosine similarity among the patch of outputs of Transformer blocks. This visualization empowers users to independently delve into the associated content.
\item \textbf{Providing a variety of interactive visualization methods to support user exploration.} To deliver an exceptional user experience, it is crucial to thoughtfully utilize diverse interactive visualization techniques. For example, when users have limited knowledge of convolutions, a combination of overview and detailed visualizations can facilitate knowledge exploration. The inclusion of animated elements can also captivate users, making it easier for them to grasp the underlying principles and implementation processes.
\item \textbf{Deploying the proposed approach as a web-based application.} In order to expand the system's accessibility to a wider audience, we have implemented the visualization tool on a web-based platform. This eliminates the necessity for any software installations, allowing users to easily access it through a web browser. This approach simplifies the process of acquiring essential knowledge about ViT and provides users with the opportunity to experience the operational procedures.
\end{itemize}

\section{Interface}\label{sec5}

The visualization system in this paper is primarily composed of four layers of visual views: Model Overview, Knowledge Background Graph, Model Detail View, and Interpretation View. An overview of the system is shown in Fig.~\ref{fig1}. The first three layers of views explain the operational flow of ViT from the perspectives of overall structure, detailed interpretation, and mathematical operations. They enable users to comprehend the underlying mathematical operations and the transformation processes between layers in the model. The fourth layer, the Interpretation View, is designed to provide in-depth understanding of the ViT model by calculating the cosine similarity between patches. Users need to click on ``Interpretation View" in the top right corner of the Model Detail View.

\begin{figure}[t]
  \centering
  \includegraphics[width=0.5\textwidth]{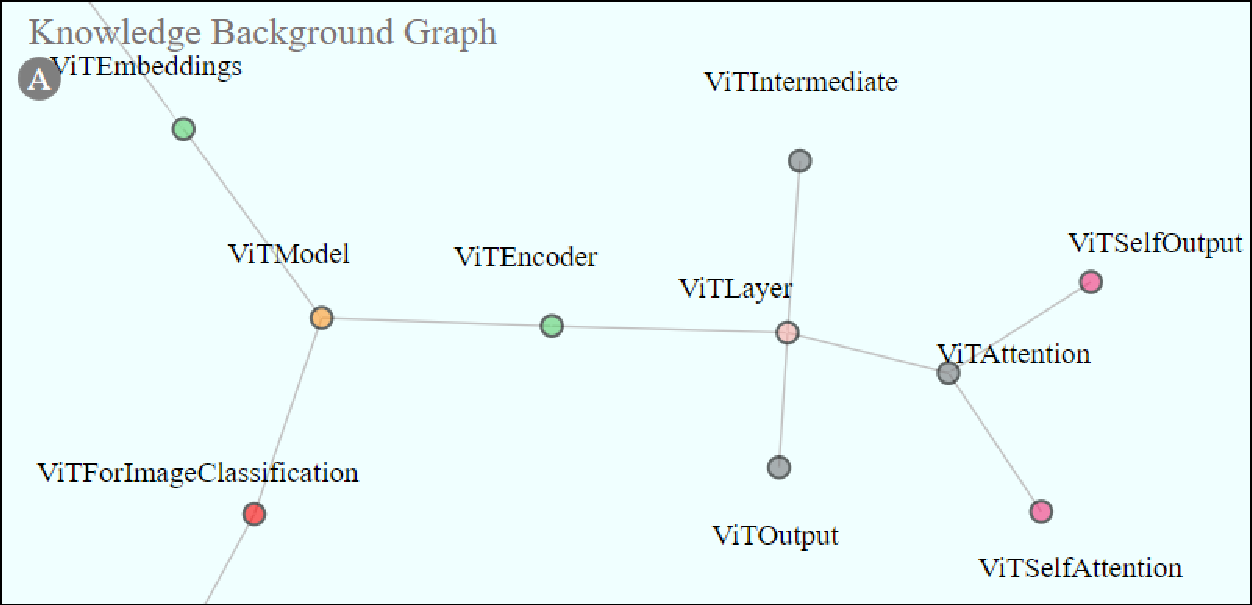}
  \includegraphics[width=0.5\textwidth]{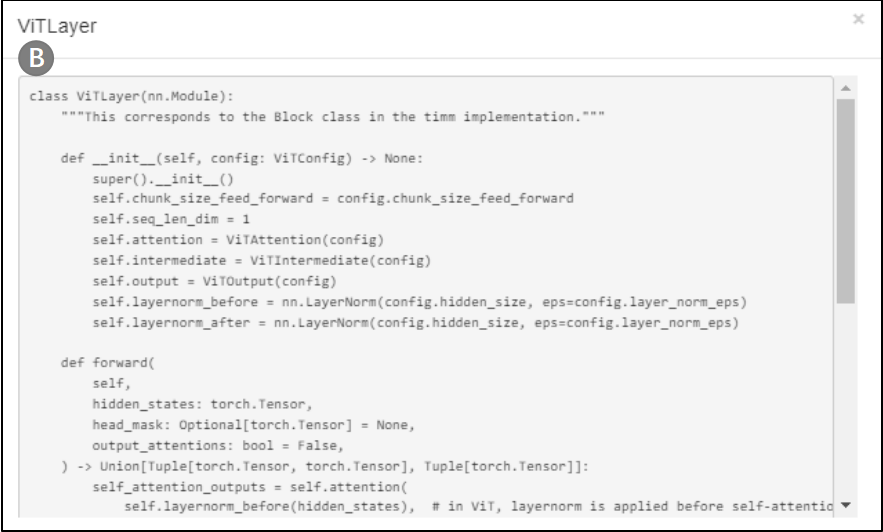}
  \caption{Knowledge Background Graph. (A) depicts as a force-directed graph, visualizing crucial concepts and the code's architecture. (B) displays the specific content of the node through modal box.}
  \label{fig4}
\end{figure}

\subsection{Model Overview}

When beginners first encounter a model, it's essential to begin with an understanding of the model's overall structure before delving into the details. Therefore, EL-VIT incorporates a Model Overview, as illustrated in Fig.~\ref{fig3}, which assists users in constructing a mental model of the model. This aids users in gaining a better understanding of the entire process, clarifying how the input from image data ultimately leads to the output classification.

For user convenience, the model overview incorporates tooltip functionality, as depicted in Fig.~\ref{fig3}(A). When a user hovers the mouse pointer over a specific intermediate step in the ViT model, a tooltip appears to provide a brief explanation of the term. This prevents users from feeling puzzled when encountering new terminology. Given the deep layer structure of the ViT model, representing the entire process in a single image could become cluttered. Therefore, the visualization system employs an interactive approach that combines focus+context. When a user clicks on ``transformer encoder", the interface displays its 12-layer structure. Clicking on one of the layers reveals the composition of the Transformer block, while clicking in a blank area will navigate the user to the previous level. The specific effects are illustrated in Fig.~\ref{fig3}(A), Fig.~\ref{fig3}(B), and Fig.~\ref{fig3}(C).

\subsection{Knowledge Background Graph}

In the process of learning about the ViT model's operation, users must not only acquire theoretical knowledge but also understand its practical implementation. Therefore, EL-VIT incorporates a Knowledge Background Graph, primarily showcasing the concepts of relevant terms and the source code used for image classification by the ViT model. The conceptual explanations of terms assist users in understanding ViT-related content, while  the  source  code aids users in comprehending the ViT model through its implementation process.

For example, the scenario where users encounter difficulties in comprehending the image classification implementation process of ViT. In such cases, users can refer to the Knowledge Background Graph within the visualization system. Since the model is organized by class in the source code, each node in this view corresponds to a specific class, and each edge can signify the presence of call relationships between nodes representing classes. The nodes within the view are generated using a force-directed layout algorithm. When users click on nodes of interest, a modal box appears, displaying the pytorch source code implementation process corresponding to that node. As illustrated in Fig.~\ref{fig4}(A), if a user wishes to understand the implementation process of ``VITLayer", they can click on the corresponding node to reveal its internal implementation details, as depicted in Fig.~\ref{fig4}(B). This process helps users gain insights into the code and its interactions with other classes.

For the purpose of enhancing the visual quality of the graphics, and to avoid any overlap between text and graphics, in the design of force-directed graphs in this paper, text information is incorporated as nodes in the force-directed graph, participating in the computation of the force-directed layout. Additionally, text node forces are introduced, which represent attractive forces between nodes and text. The combined action of multiple forces ensures that labels and actual nodes are mutually repulsive, preventing overlap, while also exerting attractive forces to bring text and nodes closer together.

\begin{figure}[t]
  \centering
  \includegraphics[width=0.5\textwidth]{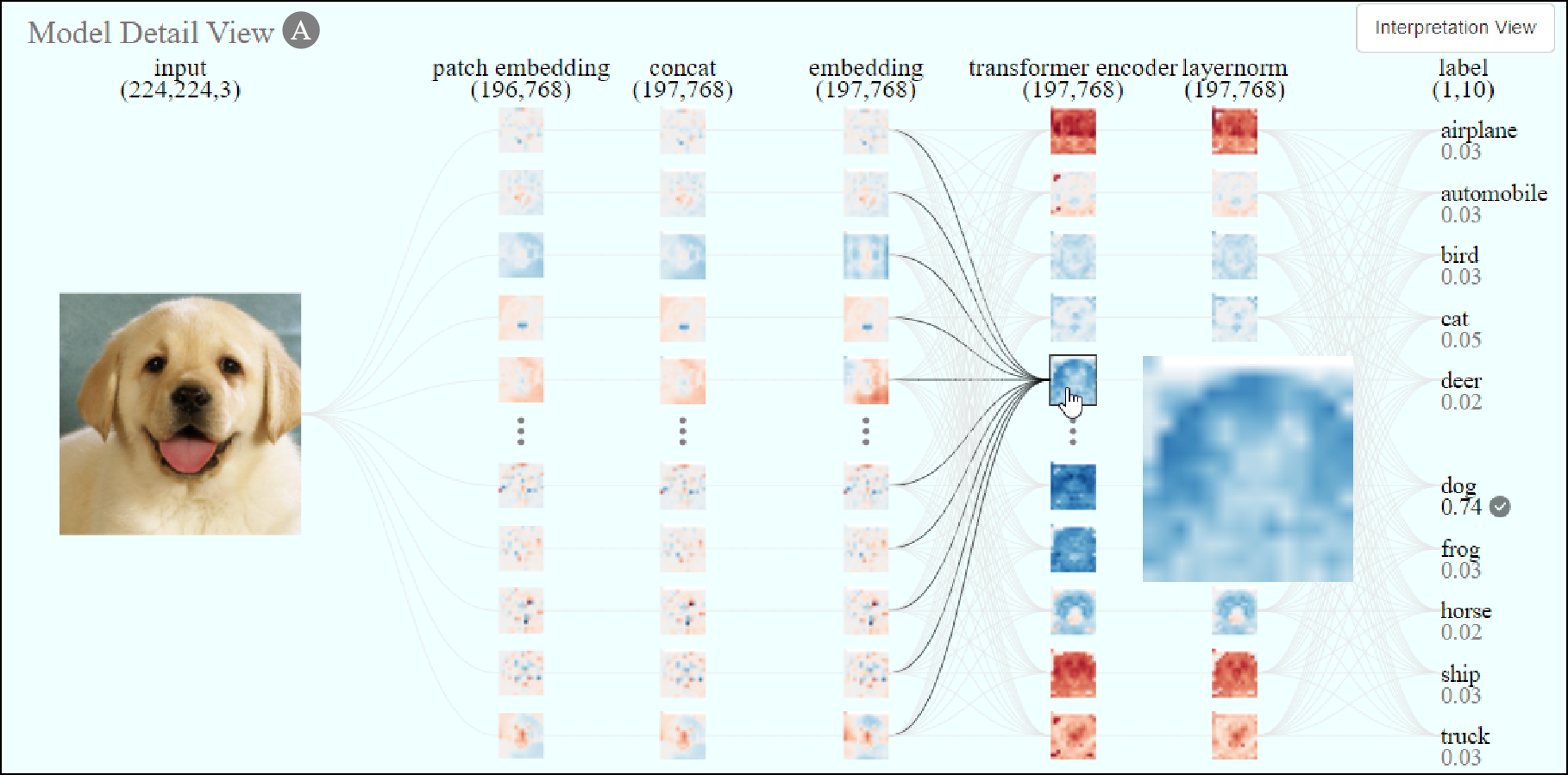}
  \includegraphics[width=0.5\textwidth]{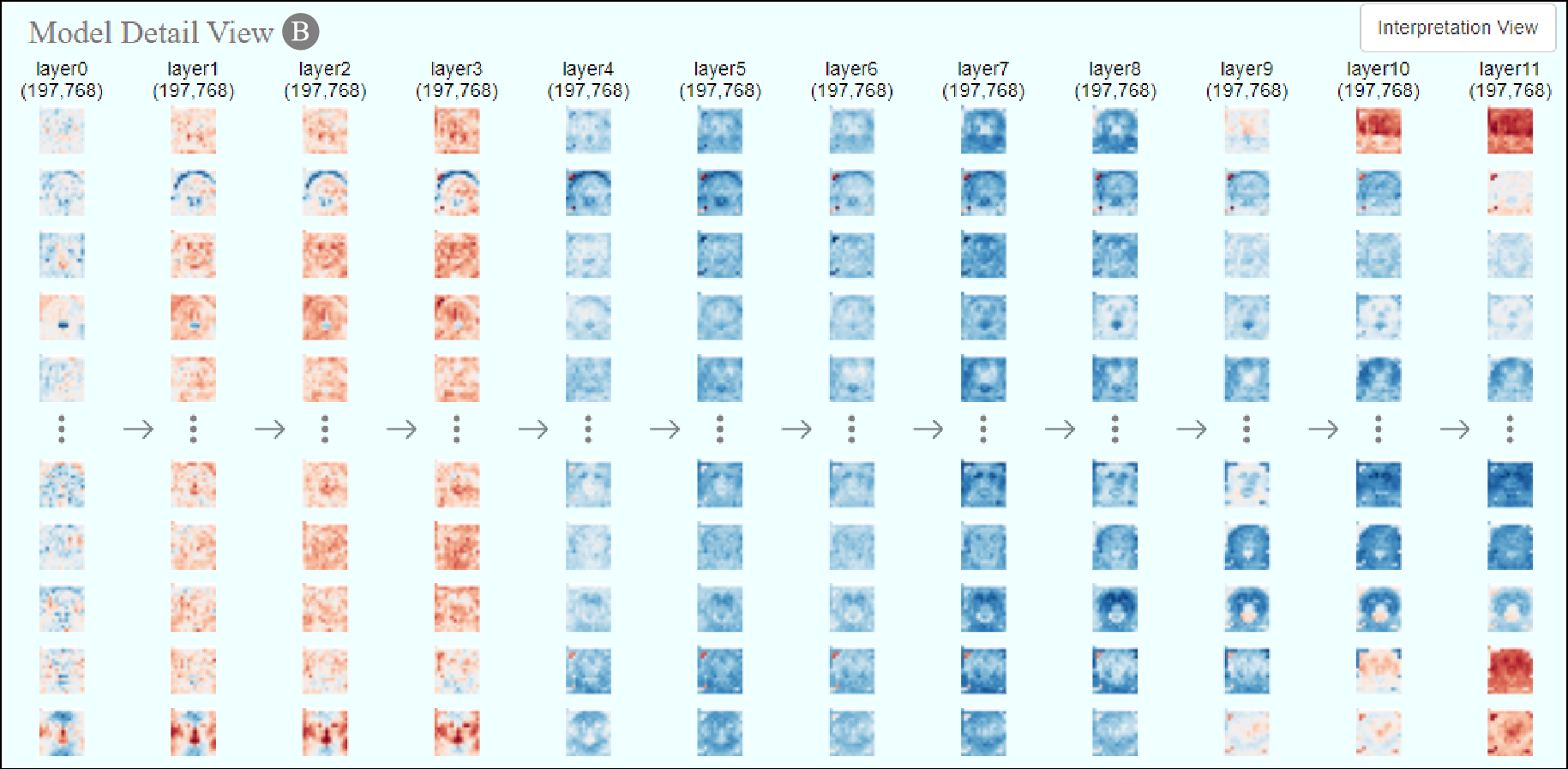}
  \includegraphics[width=0.5\textwidth]{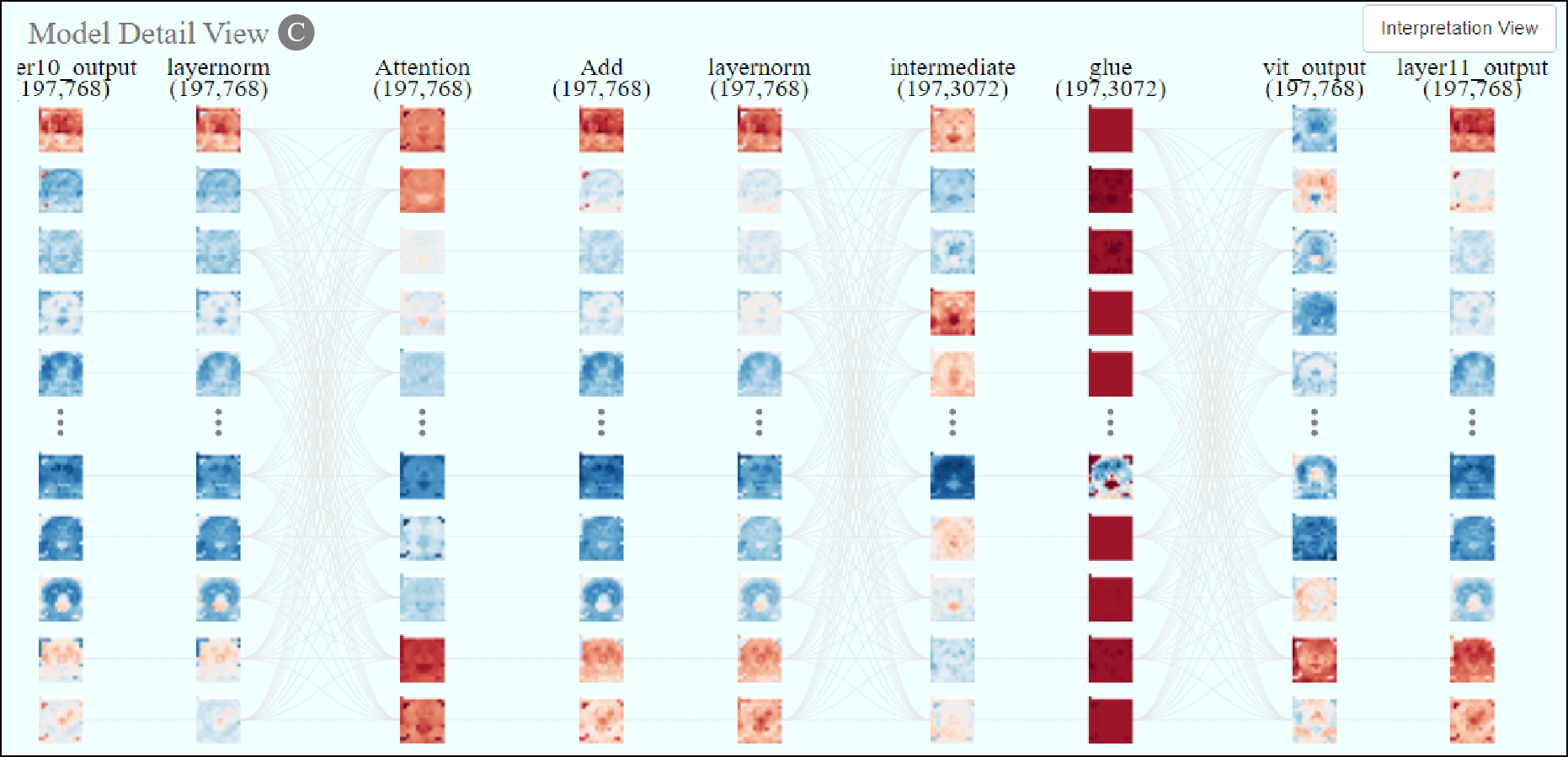}
  \includegraphics[width=0.5\textwidth]{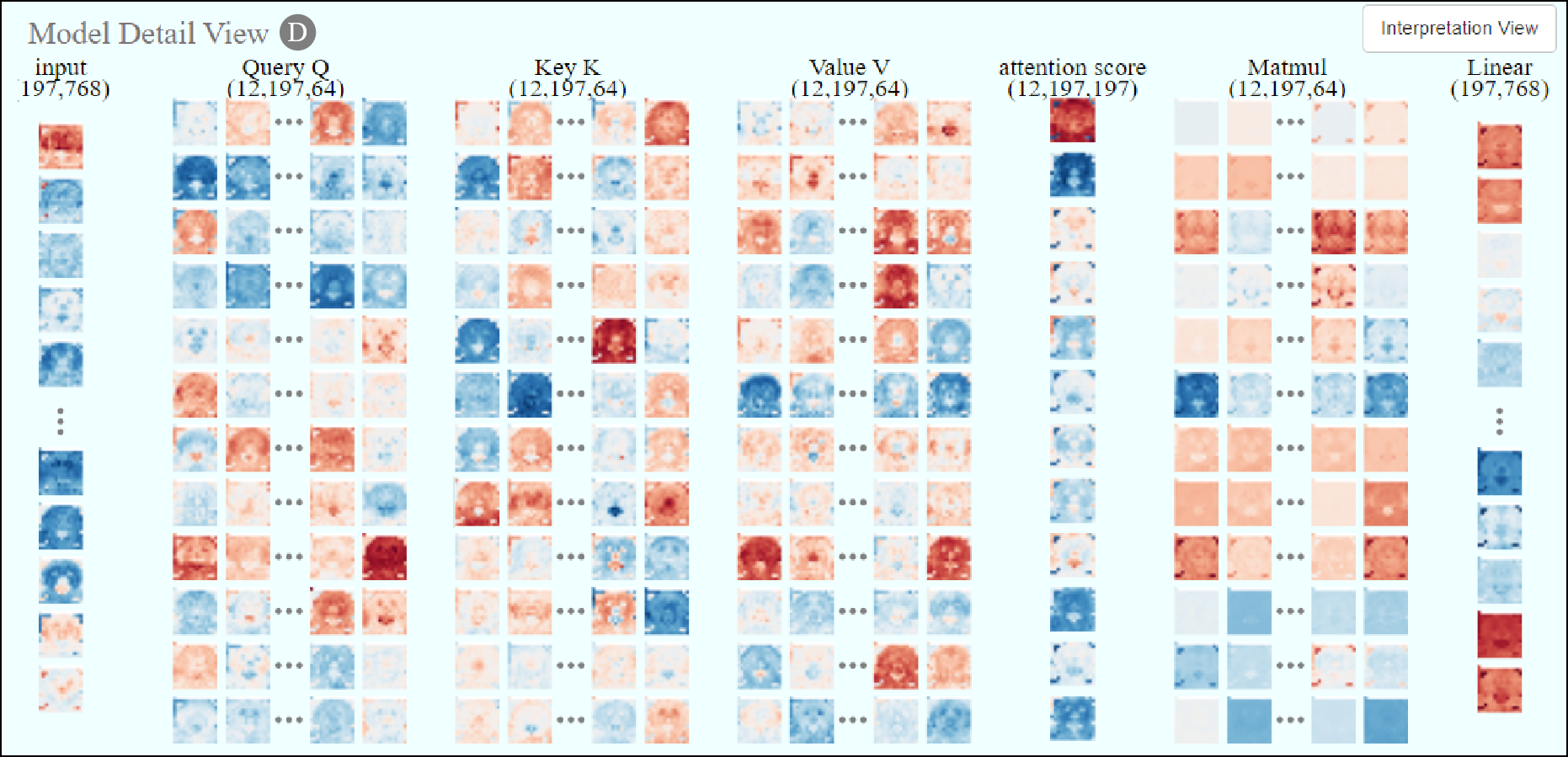}
  \caption{Model Detail View. (A) depicts the overall model process of parameters visualization. (B) displays parameter visualizations of the outputs from 12 Transformer blocks. (C) illustrates parameter visualizations for each internal layer within a Transformer block. (D) visualizes parameters of the internal layers within the Multi-Head Attention layer.}
  \label{fig5}
\end{figure}

\begin{figure}[h]
  \centering
  \includegraphics[width=0.3\textwidth]{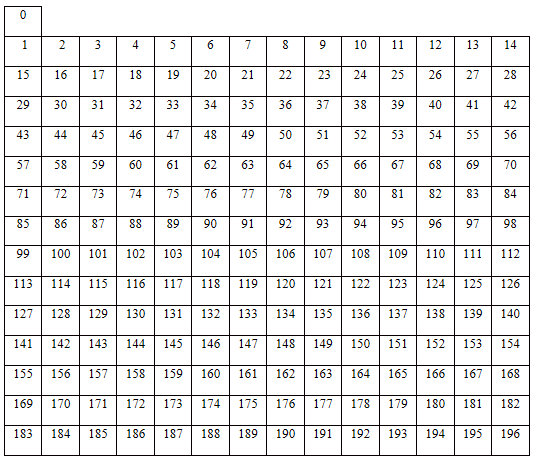}
  \caption{The resulting 2D matrix reshaped from an original vector. The number within each cell indicates the original index of each element in the original vector.}
  \label{fig6}
\end{figure}

\subsection{Model Detail View}
In the process of learning about the ViT model, beginners frequently grapple with understanding the detailed data flow, the transformations between layers, and the specific mathematical operations within each layer. Hence, EL-VIT introduces a Model Detail View.

\subsubsection{Visualization}

Fig.~\ref{fig5} demonstrates the utilization of a focus+context visualization approach in the Model Detail View, systematically guiding users through each layer. This view offers insights into the internal structure of each layer and the vector transformation processes. Users can conveniently navigate backward through the layers by clicking on empty spaces, facilitating a clear understanding of the data's transformation from the initial 224×224×3 image format to a 1×10 matrix, serving as the final classification prediction. To aid users in observing classification predictions, this paper employs iconography to highlight the class with the highest prediction probability.

To effectively visualize the graphical results, EL-VIT does not directly visualize parameters. For instance, in the embedding layer, as observed in Fig.~\ref{fig2}, a vector of size 197×768 is obtained through previous layer operations. This paper transposes the vector, resulting in a 768×197 vector, which can be perceived as 768 individual 1×197 vectors. Each of these individual vectors is reshaped into a 2D matrix, where the shape of the 2D matrix and the corresponding element locations are shown in Fig.~\ref{fig6}. Finally, the parameter values $x_i$ are subject to transformation according to Equation \ref{eq6}, normalizing them within the [0, 1] range to fulfill the input criteria for color interpolation, culminating in the effect depicted in Fig.~\ref{fig6}. In this context, $x_i$ denotes the parameter value associated with the i-th patch, range denotes the difference between the maximum and minimum values among these 197 patches, and min represents the minimum value among these 197 patches.

\begin{equation}
x_i=\frac{x_i-range}{min} \label{eq6}
\end{equation}

\begin{figure*}[htbp]
  \centering
  \includegraphics[width=0.94\textwidth]{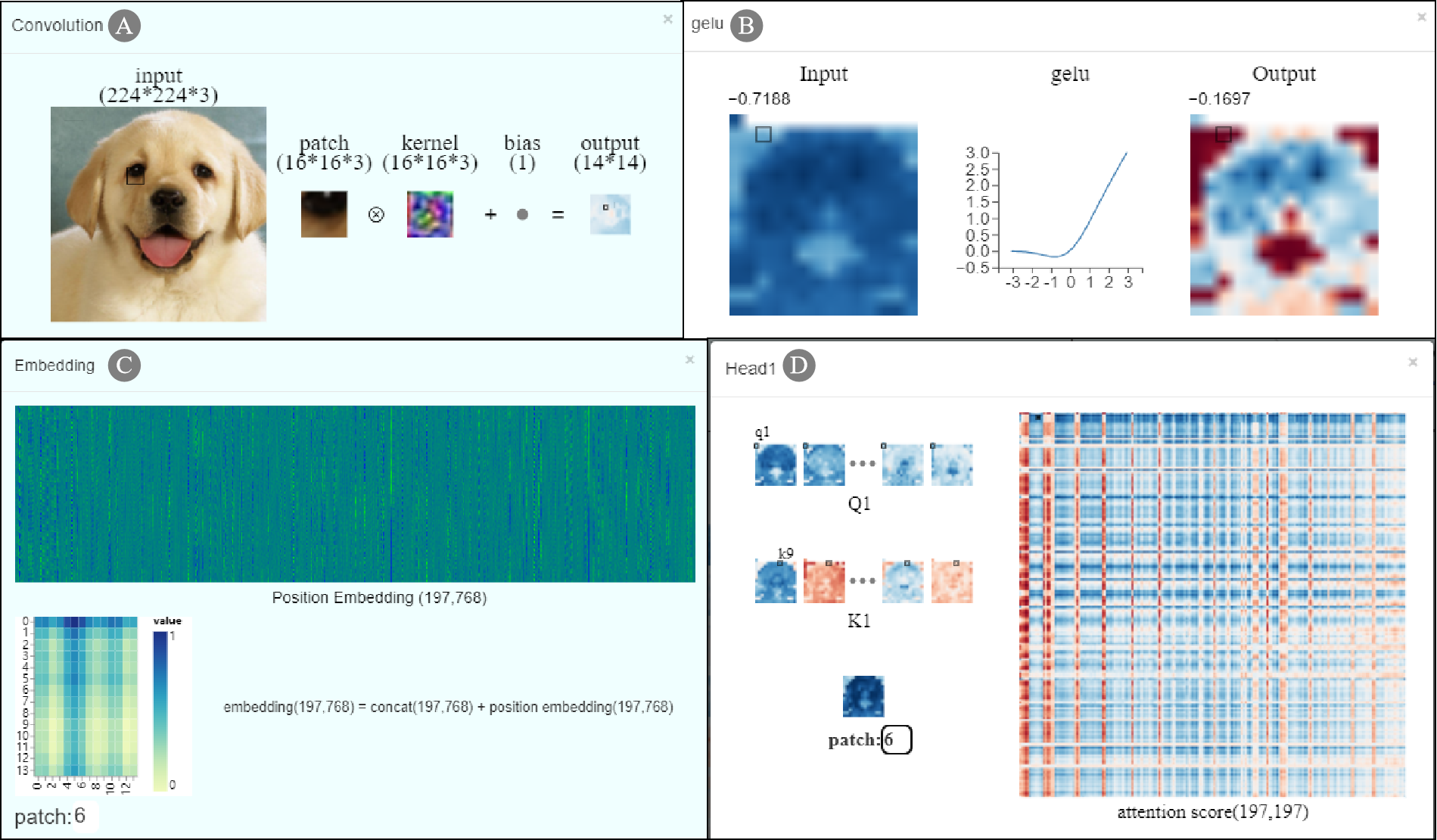}
  \includegraphics[width=0.4\textwidth]{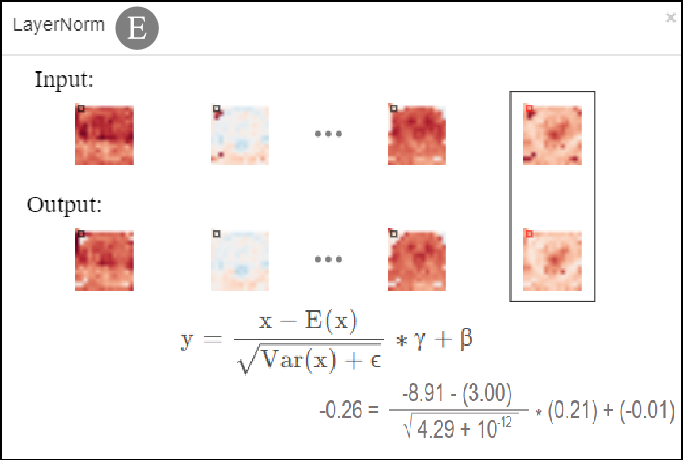}
  \caption{Interactive Views. (A) illustrates the convolution process. (B) presents the graph of the gelu function along with its input and output results. (C) depicts the embedding process while visualizing positional information. (D) shows the computation process of attention scores in the first head. (E) displays the operation of LayerNorm.}
  \label{fig7}
\end{figure*}

\subsubsection{Interactive View}

Users need to comprehend not only the overall structure of the model but also the underlying principles and mathematical operations of each layer. Therefore, this paper presents the internal workings of each layer in the form of modal boxes to enhance the interactive exploration experience for users.

For the visualization of convolution operations, as illustrated in Fig.~\ref{fig7}(A), the input image is 224×224×3, divided into 196 patches, with each patch sized at 16×16×3. Each patch is subjected to convolution using a 16×16×3 kernel and is subsequently augmented with bias terms, yielding a final 14×14 matrix. EL-VIT utilizes animation to illustrate the complete convolution procedure, demarcating the currently convolving patch with a rectangular frame for user clarity.

For the visualization of the non-linear activation function gelu, as illustrated in Fig.~\ref{fig7}(B), the view initially provides a visualization of the gelu function. When the input value is greater than 0, the gelu function essentially produces linear output, whereas when the input value is less than 0, the gelu function generates non-linear but continuous output. This view also employs animation to visualize details, with textual information accompanying the movement of the rectangular box, thus aiding users in observing data changes.

For the visualization of the embedding layer,  as illustrated in Fig.~\ref{fig2}, the embedding layer is a result of the concat layer combined with positional embedding, as demonstrated in Fig.~\ref{fig7}(C). Initially, the magnitudes of vector values are visually represented through a heatmap. To facilitate users in comprehending positional embedding more effectively, this visualization method employs a heatmap to illustrate the similarities in size between a particular patch and others. For instance, inputting ``6" in the text box will reveal the positional information for the sixth patch. Upon observation, users will notice that the colors are deeper in the row and column corresponding to this patch, aligning with their understanding of positional information.

For the visualization of attention scores, as illustrated in Fig.~\ref{fig7}(D), attention scores are presented in the form of a heatmap, positioned to the right of the modal box. Animated sequences are employed to guide users in comprehending the computation between Q and K for an individual head, thereby providing insight into the origin of the attention score matrix. This view also incorporates an exploratory feature designed to assist users in extracting specific patch information for the current head, enabling users to understand the knowledge acquired by the current patch during the forward process. For example, when the number ``6" is input into the text box, it reveals the weight information associated with the vector of the 6th patch and presents it visually in image format.

For the visualization of LayerNorm, which normalizes different features for each patch to ensure a stable distribution, as illustrated in Fig.~\ref{fig7}(E), the black small square outlines the region representing the normalization of the current patch's 768-dimensional features. It calculates the mean and variance of the current distribution, where $\gamma$ and $\beta$ denote the standard deviation and mean after normalization. Due to the high dimensionality, only a subset of features can be visually displayed in this paper. The red square frames the region representing the normalization of a specific dimension within the current patch, aligning textual information with the corresponding computational formula and moving along with the red frame. To further emphasize the visualization effect, this paper employs a large rectangular box to highlight the dimensions being computed. This view is presented through animation, aiming to engage users in the learning process.

\subsubsection{Text Detail}

\begin{figure}[h]
  \centering
  \includegraphics[width=0.25\textwidth]{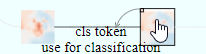}
  \caption{Text annotation visualization.}
  \label{fig8}
\end{figure}

In the case of certain basic operations, EL-VIT opts not to employ interactive views for explanations but rather employs simple text annotations for visualization, as illustrated in Fig.~\ref{fig8}. The image on the right is obtained by adding a CLS token to the left one. Users can simply hover their mouse over the image to access text annotations for explanation. Additionally, bounding boxes within the image highlight the position where CLS is added. Throughout this visualization system, several similar annotations exist to assist users in exploring information.

\subsection{Interpretation View}

EL-VIT adopts a novel perspective for model interpretation. For the Transformer encoder, it consists of 12 layers of Transformer blocks, and our focus lies on the output of these 12 Transformer blocks. As depicted in Fig.~\ref{fig5}(B), the output of any layer in the Transformer block is a vector of size 197×768. We assume that $\beta \in R^{768}$ represents an arbitrary token, such that the cosine similarity between any two tokens $\beta_i$ and $\beta_j$ is given by:

\begin{equation}
s(\beta ^i, \beta ^j)=\frac{\beta ^i\cdot  \beta ^j}{\left \| \beta ^i \right \| \cdot \left \| \beta ^j \right \| } \label{eq7}
\end{equation}

\begin{figure}[h]
  \centering
  \includegraphics[width=0.48\textwidth]{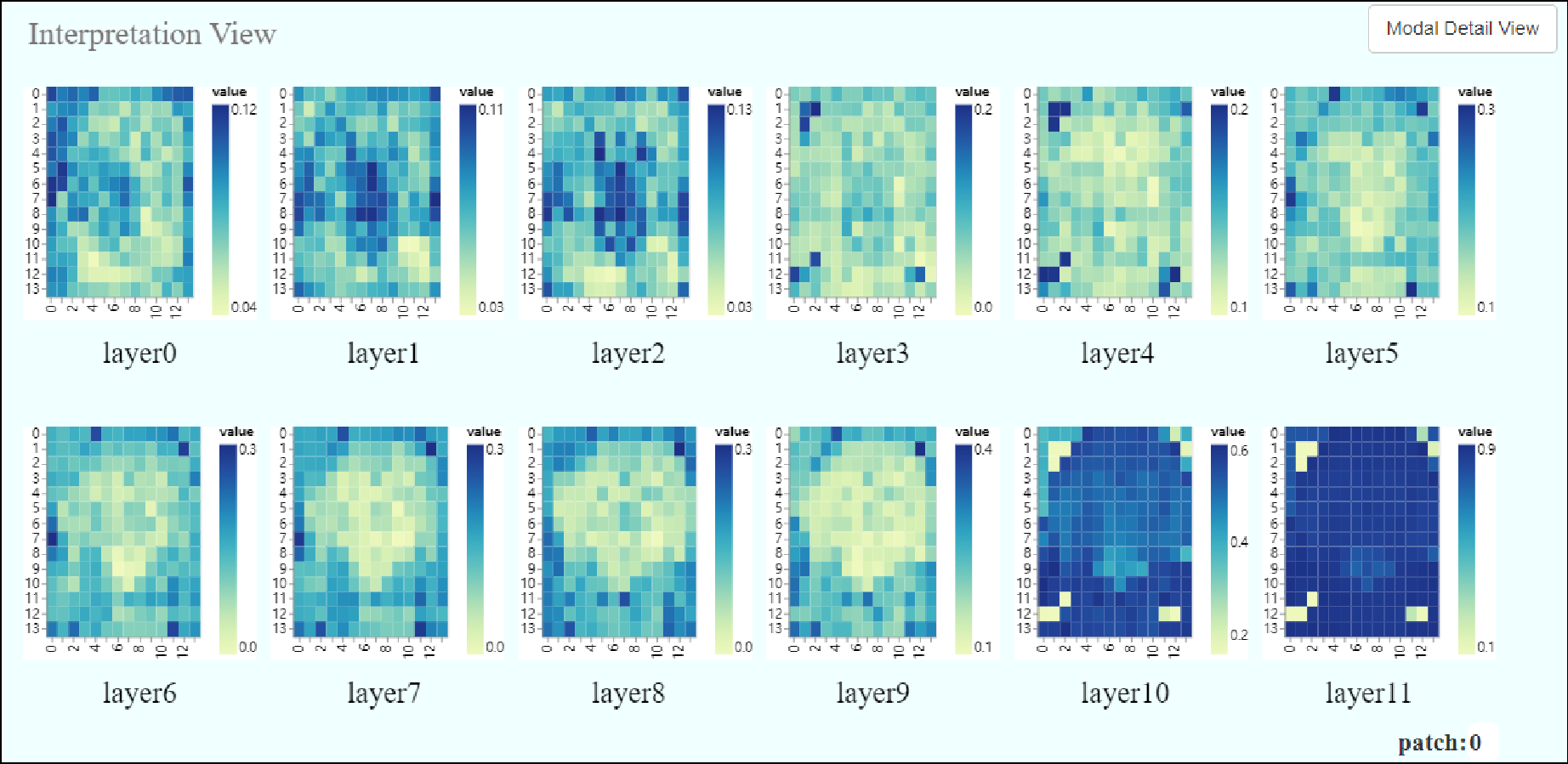}
  \caption{Interpretation View. This view allows for the alteration of text box values, facilitating the observation of the similarity relationship between a particular patch and the remaining 196 patches.}
  \label{fig9}
\end{figure}

For the original 197×768 vectors, this paper divides them into two parts: the CLS token and the remaining 196×768 vectors. Subsequently, cosine similarity calculations are performed between the CLS token and the other 196 tokens to obtain a 1×196 vector. Finally, this vector is reshaped into a 14×14 image. The specific results are shown in Fig.~\ref{fig9}. The heatmap representation visually conveys the magnitude of cosine similarity values, with bluer hues signifying greater similarity.

\subsection{Web-based Implement}

EL-VIT is a web-based implementation that does not involve any backend programming languages. We have employed TensorFlow.js\cite{b27} in conjunction with d3.js\cite{b28} to realize the visualization of the model. Users can experience ViT simply by accessing the webpage on the frontend. The model used in this paper\cite{b29} has been imported from Hugging Face and is a fine-tuned ViT model on the cifar10 dataset, employed for image classification.

\section{Usage Scenarios}\label{sec6}
\subsection{Learning the Forward Process}

\begin{figure}[t]
  \centering
  \includegraphics[width=0.5\textwidth]{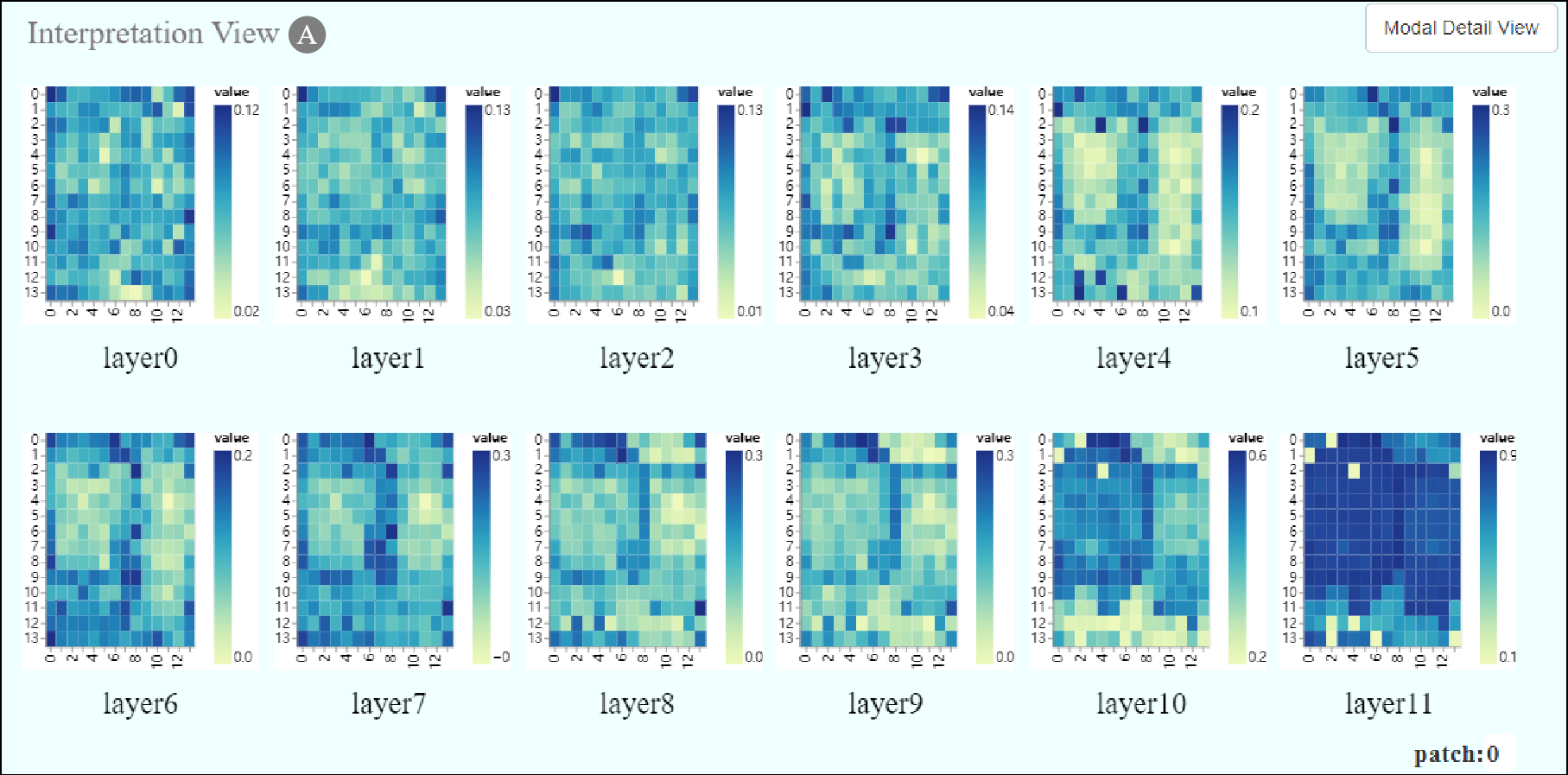}
  \includegraphics[width=0.5\textwidth]{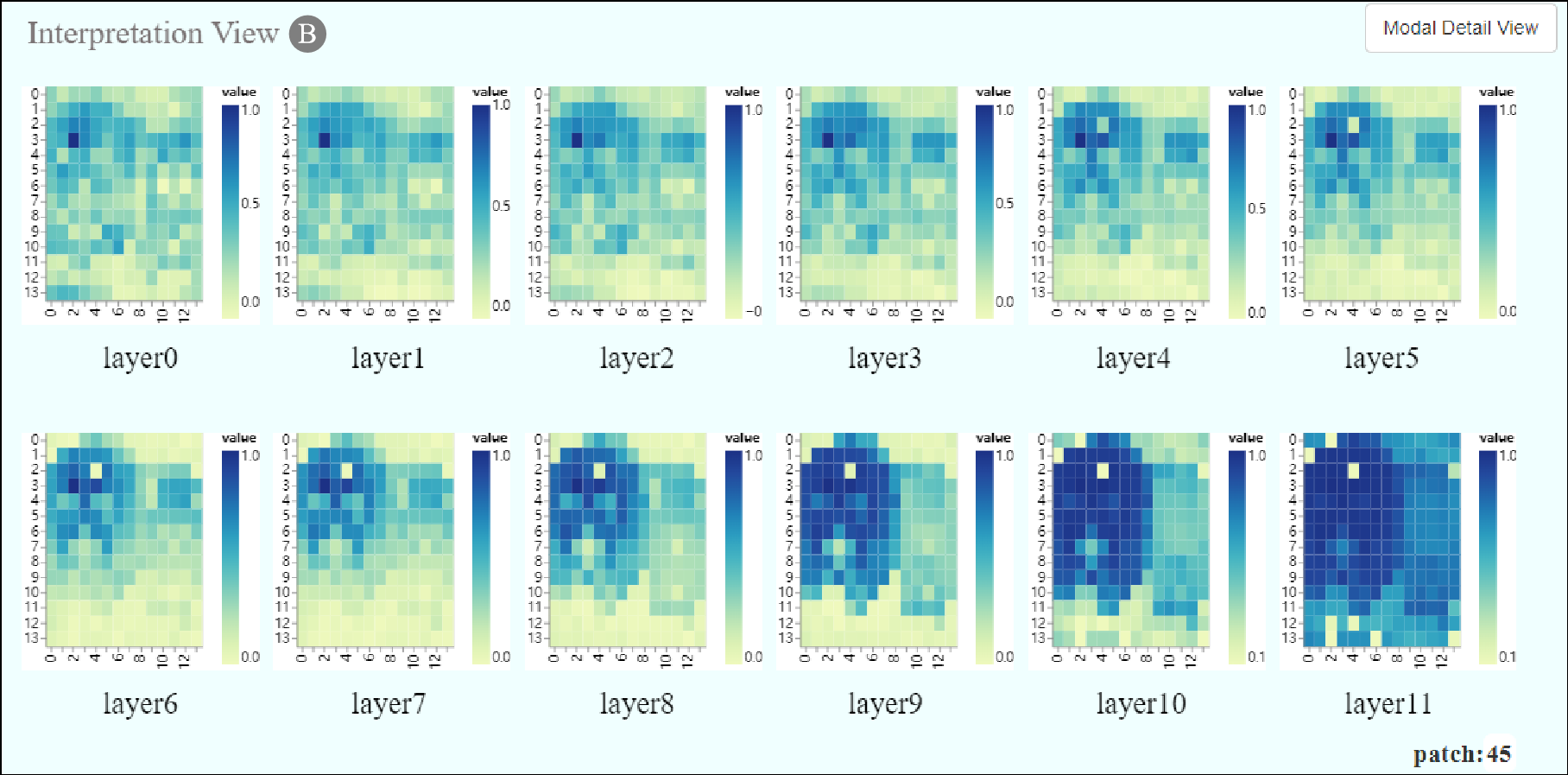}
  \includegraphics[width=0.5\textwidth]{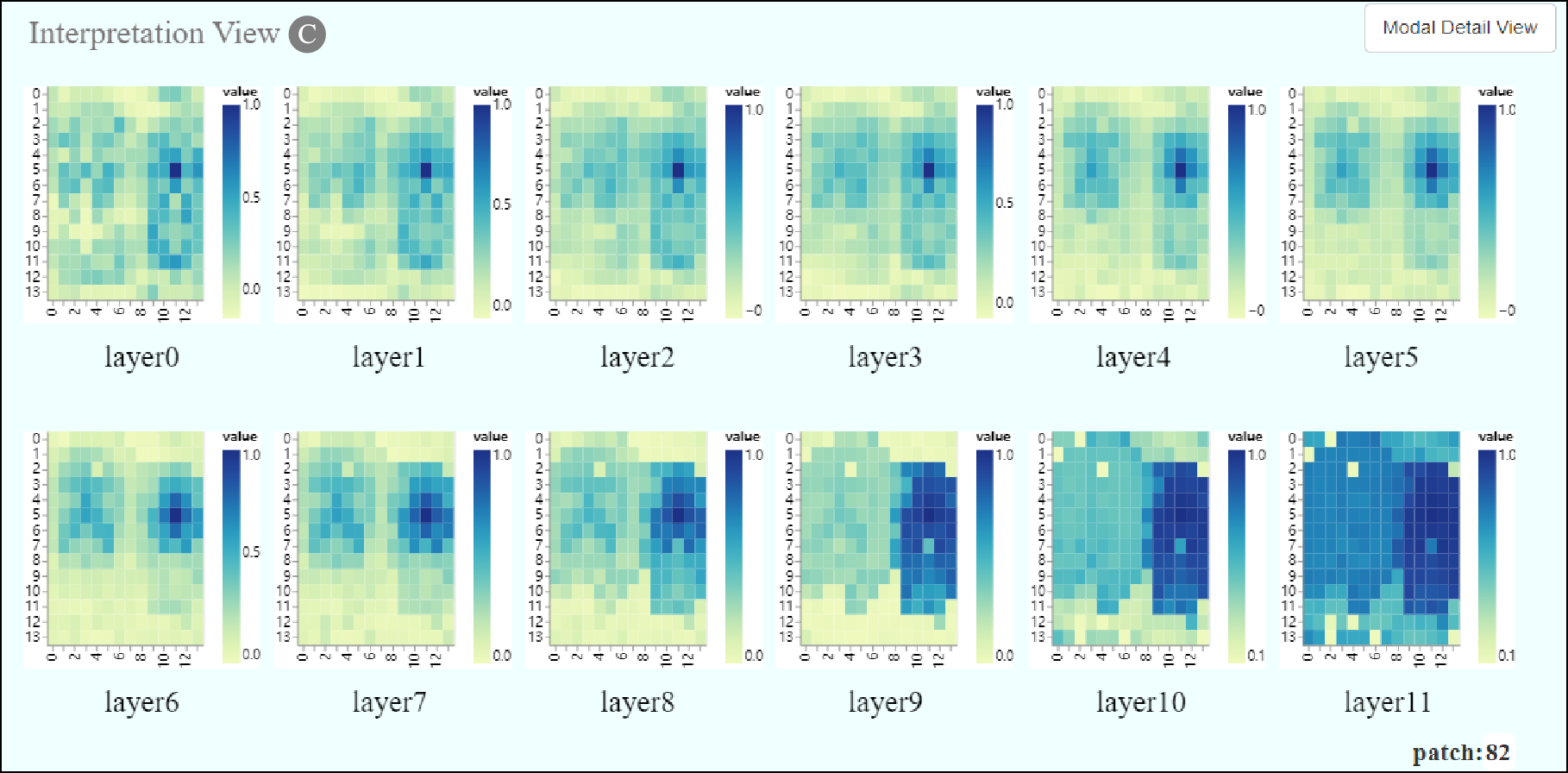}
  \caption{Similarity view of patch. (A) illustrates the similarity map between the first patch (CLS) and the 196 patches related to the image. (B) displays the similarity map between the 45th patch (patch corresponding to the dog) and the 196 patches related to the image. (C) shows the similarity map between the 82nd patch (patch corresponding to the cat) and the 196 patches related to the image.}
  \label{fig10}
\end{figure}

When computer vision students, who are learning about image classification, discover that the classification performance of ViT is superior to CNN, they decide to enhance the accuracy of image classification by learning and applying ViT since they are not very familiar with the ViT model. Consequently, they embark on exploring EL-VIT.

Firstly, the students begin their exploration of the visualization system by starting with the model overview, as shown in Fig.~\ref{fig3}. This initial step provides them with a general understanding of the entire operational process of ViT, helping them establish a mental model. Subsequently, the students seek to delve deeper into the details, encountering difficulties while exploring the model detail view. They are uncertain about why a 224×224×3 image data is transformed into 196×768 after patch embedding. By clicking on the corresponding image to reveal a modal box, as illustrated in Fig.~\ref{fig7}(A), they swiftly comprehend that this is a convolutional process. Through animations and their foundational knowledge, they efficiently calculate the output dimensions to be precisely 196×768, aligning with the intended content of the visualization. Continuing their exploration through the subsequent layers, the students ultimately gain insight into the step-by-step calculation of classification predictions. Due to the model's complexity, the students face challenges during code replication. Consequently, they delve into the knowledge background graph to understand the code's construction, as shown in Fig.~\ref{fig4}. For instance, if a user aims to create ``VITModel", it consists of components such as ``VITEncoder" and ``VITEmbedding", with each layer housing various sub-layers. The students find that this interactive tool makes exploring ViT more engaging than traditional learning methods like lectures or video tutorials.

\subsection{Interpreting the Workings}

In terms of Transformer interpretability, in most cases, users analyzed attention weights. Take the attention scores in Fig.~\ref{fig7}(D) as an example. For a 197×197 vector, this paper extracts the CLS (0th dimension) 1×197 vector, then reshapes the subsequent 196 parameters into a 14×14 image. However, this approach may not always be effective. Therefore, users can activate the interpretation view, as illustrated in Fig.~\ref{fig9}. For a single entity image, users can clearly see that the patch corresponding to the dog has relatively dark colors, indicating that CLS is very similar to the patch corresponding to the dog and dissimilar to the background. For multiple entities, as shown in Fig.~\ref{fig10}(A), CLS is similar to the patches corresponding to both the cat and the dog, resulting in high probabilities for both cat and dog outputs.

As shown in Fig.~\ref{fig10}(B), when we changed the patch in the lower right corner to 45, corresponding to the dog, all patches associated with the dog become quite similar. Similarly, when we replaced the patch with the one corresponding to the cat, there was a corresponding effect, as shown in Fig.~\ref{fig10}(C). This observation aligns well with the findings of Nguyen et al\cite{b30}. We believed that it was precisely the spatial similarity among patches of the same object that leads to dimension reduction and clustering effects. Moreover, by closely examining the data, we hypothesized that shallow-layer patches interact with the surrounding patches, resulting in high similarity among patches corresponding to the cat in the vicinity. It is only in the deeper layers that all patches corresponding to the cat become similar.

\begin{figure}[h]
  \centering
  \includegraphics[width=0.5\textwidth]{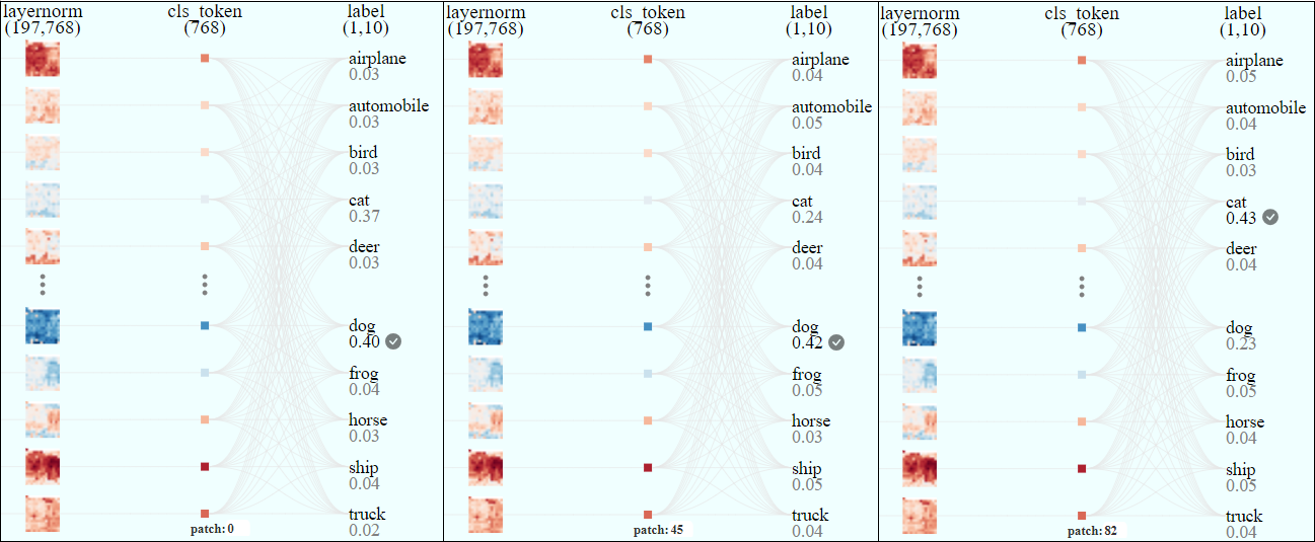}
  \caption{The left, middle, and right figures represent the calculated prediction probability maps for the 0th patch (CLS), the 45th patch (corresponding to the dog), and the 82nd patch (associated with the cat), respectively.}
  \label{fig11}
\end{figure}

To further explore, we opened the model detail view and clicked on the label tags to reveal hidden views, as shown in Fig.~\ref{fig11}. The left, middle, and right panels represent the probabilities of various classes computed for the 0th patch (CLS), the 45th patch (patch corresponding to the dog object), and the 82nd patch (patch corresponding to the cat object), respectively. We observed that when using the patch associated with the dog object instead of CLS for the final classification prediction, the probability of predicting as a dog increased, while the probability of predicting as a cat decreased. Similarly, when using the patch corresponding to the cat object instead of CLS for the final classification prediction, the probability of predicting as a cat increased, while the probability of predicting as a dog decreased.

\section{Limitations and Future Work}\label{sec7}

\textbf{Lacking of training process and backpropagation.} In the process of learning deep learning models, backpropagation also plays a vital role. However, in EL-VIT, all the parameters are imported, and the lack of a backpropagation process makes it challenging for beginners to understand how these parameters are derived. In the future, we plan to enrich the content of EL-VIT and gradually visualize the backpropagation process.

\textbf{Extending visualization form.} Due to the high dimensionality and large number of parameters in ViT, it is challenging to visualize all the details adequately. To address this, we aim to extend our visualization methods in the future by incorporating greater interactivity. This approach will better assist users in exploring and gaining detailed insights into specific aspects of interest.

\textbf{Further exploring the interpretability of the model.} While we have explored the interpretability of the model from another perspective, whether through the model's outputs or by investigating its interpretability through attention weights, it is essential to recognize that neither approach can provide a complete explanation of the model. Both of these methods only represent a small portion of the entire model. We hope that in the future, further research will enable us to comprehensively understand these black-box models.

\textbf{Evaluation of the educational effectiveness.} EL-VIT provides interpretability explanations and serves as an educational tool, aiding both experts and novices in knowledge acquisition. Future plans include incorporating surveys to further explore the learning experiences of beginners.

\section{Conclusion}\label{sec8}

In this paper, we introduce a user-friendly interactive visualization tool that employs intuitive visualization techniques to assist users in exploring ViT models. It caters to both experts and beginners, facilitating the acquisition of necessary knowledge. By employing cosine similarity on the output of each layer of the Transformer, we observe that patches corresponding to the same object exhibit higher similarity, and the CLS token also shares high similarity with patches corresponding to objects. Finally, we demonstrate the effectiveness of the visualization system through two usage scenarios.

\section*{Acknowledgment}

We thank anonymous reviewers for their valuable comments. This research was supported by Lee Kong Chian Fellowship awarded to Dr. Yong Wang by SMU. The demonstration video for this visualization system is located at https://youtu.be/MUmk0Fi0JKk.

\vspace{12pt}

\end{document}